\documentclass[conference]{IEEEtran}
\IEEEoverridecommandlockouts
\usepackage{cite}
\usepackage{amsmath,amssymb,amsfonts}
\usepackage{algorithmic}
\usepackage{graphicx}
\usepackage{textcomp}
\usepackage{xcolor}

\usepackage{subcaption}
\usepackage{hyperref}
\usepackage{booktabs}
\usepackage{multirow}

\def\BibTeX{{\rm B\kern-.05em{\sc i\kern-.025em b}\kern-.08em
    T\kern-.1667em\lower.7ex\hbox{E}\kern-.125emX}}
\begin{document}

\title{Color Space Learning for Cross-Color Person Re-Identification}

\author{\IEEEauthorblockN{1\textsuperscript{st} Jiahao Nie}
\IEEEauthorblockA{\textit{Interdisciplinary Graduate Programme} \\
\textit{Nanyang Technological University} \\
Singapore \\
jiahao007@e.ntu.edu.sg}
\and
\IEEEauthorblockN{2\textsuperscript{nd} Shan Lin$^{*}$}\thanks{* Corresponding author}
\IEEEauthorblockA{\textit{Rapid-Rich Object Search (ROSE) Lab} \\
\textit{Nanyang Technological University} \\
Singapore \\
shan.lin@ntu.edu.sg}
\and
\IEEEauthorblockN{3\textsuperscript{rd} Alex C. Kot}
\IEEEauthorblockA{\textit{Rapid-Rich Object Search (ROSE) Lab} \\
\textit{Nanyang Technological University} \\
Singapore \\
eackot@ntu.edu.sg}
}

\maketitle

\begin{abstract}
The primary color profile of the same identity is assumed to remain consistent in typical Person Re-identification (Person ReID) tasks. However, this assumption may be invalid in real-world situations and images hold variant color profiles, because of cross-modality cameras or identity with different clothing. To address this issue, we propose Color Space Learning (CSL) for those Cross-Color Person ReID problems. Specifically, CSL guides the model to be less color-sensitive with two modules: Image-level Color-Augmentation and Pixel-level Color-Transformation. The first module increases the color diversity of the inputs and guides the model to focus more on the non-color information. The second module projects every pixel of input images onto a new color space. In addition, we introduce a new Person ReID benchmark across RGB and Infrared modalities, NTU-Corridor, which is the first with privacy agreements from all participants. To evaluate the effectiveness and robustness of our proposed CSL, we evaluate it on several Cross-Color Person ReID benchmarks. Our method surpasses the state-of-the-art methods consistently. The code and benchmark are available at: \href{https://github.com/niejiahao1998/CSL}{\textit{\textcolor{blue}{https://github.com/niejiahao1998/CSL}}}
\end{abstract}


\section{Introduction}
\label{sec:intro}

Person Re-identification (Person ReID) is to search the identity of interest across multiple cameras under various environments and different time instances~\cite{zheng2016person,ye2021deep}. Nevertheless, typical Person ReID only considers the cross-camera challenge, in which identity images are captured across different views and backgrounds and remain in the same color profile. Despite existing methods achieving great success under color-consistent circumstances, the same identity may hold variant color profiles in real-world situations.

To propose a unified method for the above challenges, we aim at Cross-Color Person ReID tasks simultaneously, encompassing Visible-Infrared Person ReID (\textbf{VI-ReID})~\cite{wang2019beyond} and Cloth-Changing Person ReID (\textbf{CC-ReID})~\cite{fan2020learning}. The two tasks are challenged by modality gaps and variations in clothing colors, and they always occur together in the real world.

\begin{figure}[h]
    \centering
    \includegraphics[width=0.8\linewidth]{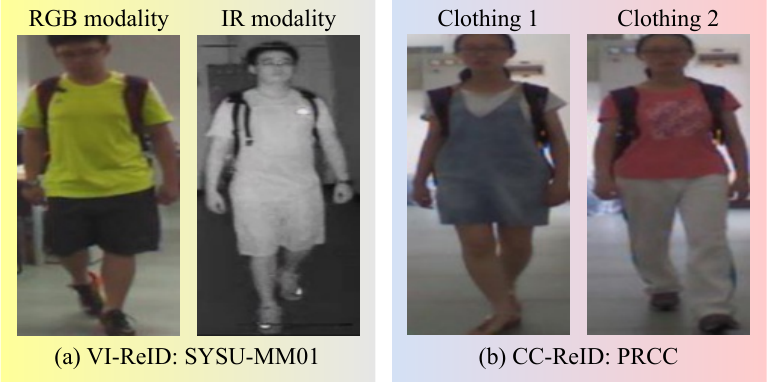}
    \caption{Sample images of the same identity from SYSU-MM01~\cite{wu2017rgb} and PRCC~\cite{yang2019person} datasets. (a) VI-ReID challenge is the color difference for the entire image. (b) CC-ReID challenge is the color difference for the clothing region only. Best viewed in color.}
    \label{fig:sample}
    \vspace{-4mm}
\end{figure}

Infrared and RGB cameras are adopted together to capture images during the night-time or under poor-light environments, which forms the \textbf{VI-ReID} task~\cite{wu2017rgb}. The main challenge of VI-ReID is the significant modality discrepancy between RGB and Infrared modalities. Existing methods can be broadly categorized into two directions: \textit{(i)} input image operations, such as image generation~\cite{wang2019rgb,wang2020cross}; and \textit{(ii)} feature space projections~\cite{ye2021deep,park2021learning}. Nevertheless, image generation introduces computation costs and noise, and simple feature projection is incapable of mitigating the significant gap between RGB and Infrared modalities.

The \textbf{CC-ReID} is to search the same identity wearing different clothing sets. Most existing methods use auxiliary information (\textit{e.g.} \textit{sketch}~\cite{yang2019person} and \textit{pose information}~\cite{qian2020long}) to avoid interference from clothing changes. However, replacing the original images with generated auxiliary information leads to information loss and introduces noise. Consequently, recent approaches turn to using RGB modality only instead~\cite{gu2022clothes}.

For VI-ReID, the major discrepancy between RGB and Infrared modalities is the color difference of the entire images, as shown in Fig.~\ref{fig:sample}(a). For CC-ReID, color is still one significant difference, which is confined to the clothing area, as shown in Fig.~\ref{fig:sample}(b). Therefore, we believe that one unified method can solve these two tasks simultaneously. It could be a very important research direction because the above two problems always occur together in real-world surveillance systems, and we encompass them into \textbf{Cross-Color Person ReID} tasks. With this insight, we propose \textbf{C}olor \textbf{S}pace \textbf{L}earning (\textbf{CSL}) to guide the model focus on non-color information. We first design an Image-Level Color-Augmentation (ICA) module to change the color of inputs and mix up the generated image with the original, reducing the color change artifacts. The augmented inputs increase color diversities while keeping most texture information. Then we design a Pixel-Level Color-Transformation Module (PCT), which learns the relation between color channels and projects the input images into a new color style for color-insensitive feature matching. Extensive experiments demonstrate that CSL outperforms most state-of-the-art methods.

\begin{figure}[ht]
  \centering
  \vspace{-3mm}
  \includegraphics[width=0.9\linewidth]{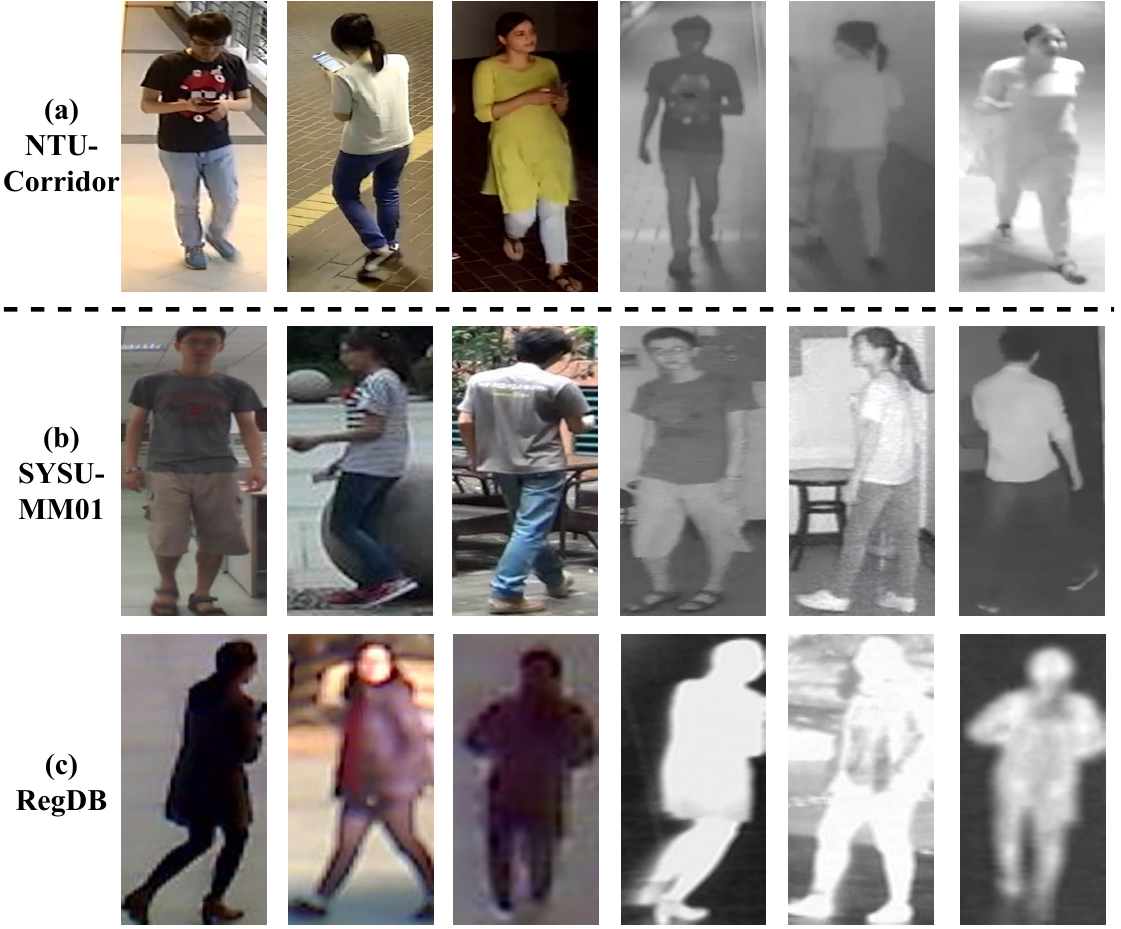}
  \caption{Sample images of our NTU-Corridor and other VI-ReID datasets. Ours are captured from the top-down surveillance view from NIR cameras.}
  \vspace{-2mm}
  \label{fig:dataset}
\end{figure}

In addition, we also construct a new VI-ReID benchmark: \textbf{NTU-Corridor}, which is shown in Fig.~\ref{fig:dataset}. We use 136 real-world surveillance cameras to capture identities on the NTU campus, which covers various environments and lighting conditions. To the best of our knowledge, our benchmark is the one closest to the real-world application scenarios because images are captured in a top-down view. It is worth mentioning that our benchmark is released with \textbf{privacy agreements} from all participants.

In summary, the main contributions of this work are as follows: \textit{(i)} We propose the Color Space Learning (CSL) method, which solves the prevalent Person ReID tasks under variant color profiles; \textit{(ii)} We establish the NTU-Corridor benchmark for VI-ReID, which is recorded in the NTU buildings by 136 surveillance cameras. Our benchmark is closest to real-world surveillance scenarios and comes with privacy agreements from all participants; \textit{(iii)} Our proposed CSL surpasses the state-of-the-arts on extensive Cross-Color Person ReID benchmarks. To the best of our knowledge, our CSL is the first method that can solve VI-ReID and CC-ReID simultaneously.

\section{Related Work}
\label{sec:relatedwork}

\noindent{\textbf{Visible-Infrared Person ReID (VI-ReID)}} is a cross-modality problem, which was proposed to solve surveillance tasks under poor-illuminance conditions. Wu~\textit{et al.}~\cite{wu2017rgb} first defined the VI-ReID problem and they directly utilized gray-scale images to bridge the gap. To mitigate the cross-modality discrepancy, some methods attempted to generate the counterpart modality by Generative Adversarial Networks (GANs)~\cite{wang2019rgb,wang2020cross,wei2021syncretic}. Nevertheless, these self-generated methods paid more attention to RGB modalities, thus Hu~\textit{et al.}~\cite{hu2022adversarial} used domain and identity information respectively to train a network. To avoid increasing computation costs, Ye~\textit{et al.} \cite{ye2021deep} further designed a two-stream ResNet50~\cite{he2016deep} backbone without generating auxiliary information. Then, Park~\textit{et al.}~\cite{park2021learning} and Ye~\textit{et al.}~\cite{ye2021channel} introduced dense cross-modality correspondences and data augmentation to alleviate the discrepancy respectively. However, existing methods mainly focused on the image-level gap and are biased on RGB modality.

\noindent{\textbf{Cloth-Changing Person ReID (CC-ReID)}} is a crucial problem under long-term target-identity search circumstances. Yang~\textit{et al.}~\cite{yang2019person} constructed the first CC-ReID dataset, and proposed a deep contour-sketch-based network for targets in different clothing sets. Cui~\textit{et al.}~\cite{cui2023dcr} disentangled
cloth-irrelevant features and cloth-relevant features and focused on the former one. Then, Qian~\textit{et al.}~\cite{qian2020long} designed a shape-embedding model to extract biological structural features, and focused on identity-relevant features. To further utilize the color-irrelevant information, Chen~\textit{et al.}~\cite{chen2021learning} reconstructed 3D features to
enhance the discriminative ability. However, the auxiliary information introduced some redundant and noisy features, as well as more computation costs.

These two problems always occur simultaneously in application scenarios, but previous methods are separately designed. We find that the common challenge for both problems is the color difference. Therefore, our method focuses on learning the color-insensitive features and aims to solve two problems at one time.
 
\begin{figure*}[ht]
    \centering
    \vspace{-1mm}
    \includegraphics[width=0.86
    \linewidth]{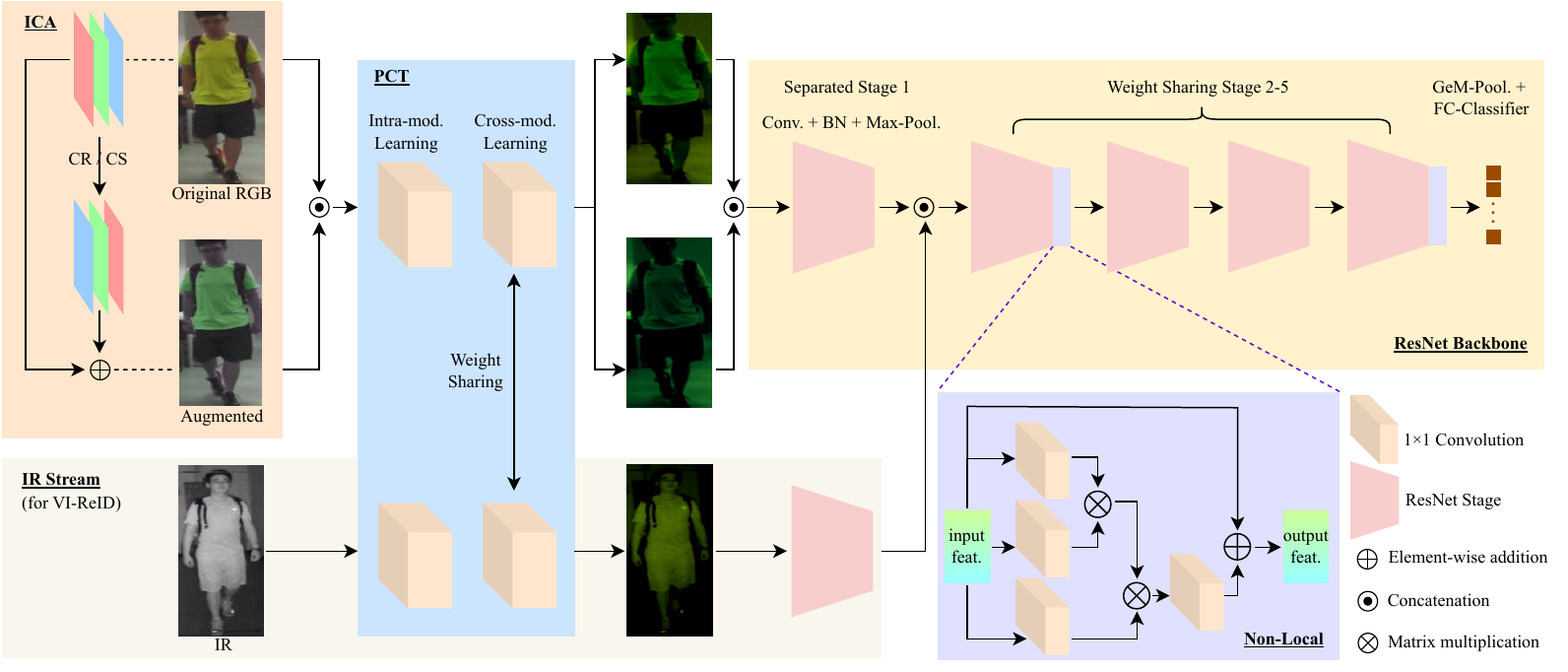}
    \vspace{-1mm}
    \caption{The framework of the proposed Color Space Learning (CSL) for Cross-Color Person ReID. It contains Image-Level Color-Augmentation (ICA) and Pixel-Level Color-Transform (PCT). The lower IR-Stream is only only used for VI-ReID.}
    \label{fig:structure}
    \vspace{-2mm}
\end{figure*}

\section{Proposed Method}
\label{sec:method}

As shown in Fig.~\ref{fig:structure}, our proposed Color Space Learning (CSL) consists of two major components, and we will organize this section as follows: \textit{(i)} Image-Level Color-Augmentation (ICA), \textit{(ii)} Pixel-Level Color-Transformation (PCT), and \textit{(iii)} Backbone and Loss function.

\subsection{Image-Level Color-Augmentation}
\label{ssec:image-level}
\vspace{-1mm}
\noindent For color-inconsistent tasks, image-level augmentation in the input stage is straightforward to mimic color changes. Our goal is to design a light and simple method, which is plug-and-play and adaptive with end-to-end structures. At the beginning, Channel Replacement Augmentation (CR)~\cite{ye2021channel} is proposed, which randomly adopts one channel to replace the other channels, and the generated images can bridge the modality gap. We denote the training set as $X=(X_V, X_I)$. Specifically, $X_V=\{\mathbf{x}_{i}^{V}\,|\,i=1,2,\cdots,N^V\}$ represents the RGB images, and $X_I=\{\mathbf{x}_{i}^{I}\,|\,i=1,2,\cdots,N^I\}$ represents the IR images. There are three channels, R, G, and B for each RGB image, \textit{i.e.} $\mathbf{x}_{i}^{V}=\{x_{i}^{R},x_{i}^{G},x_{i}^{B}\}$. The generated images from data augmentation are denoted as $\mathbf{\tilde{x}}_{i}^{V}$, and three channels are still inside, \textit{i.e.} $\{x_{i}^{C_1},x_{i}^{C_2},x_{i}^{C_3}\}$. The CR can be formulated as:
\begin{equation}
  \mathbf{\tilde{x}}_{i}^{V}=\{x_{i}^{C},x_{i}^{C},x_{i}^{C}\}, \quad C\in\{R,G,B\}.
\label{eq:ica1}
\end{equation}

\begin{figure}[ht]
  \centering
  \includegraphics[width=0.9\linewidth]{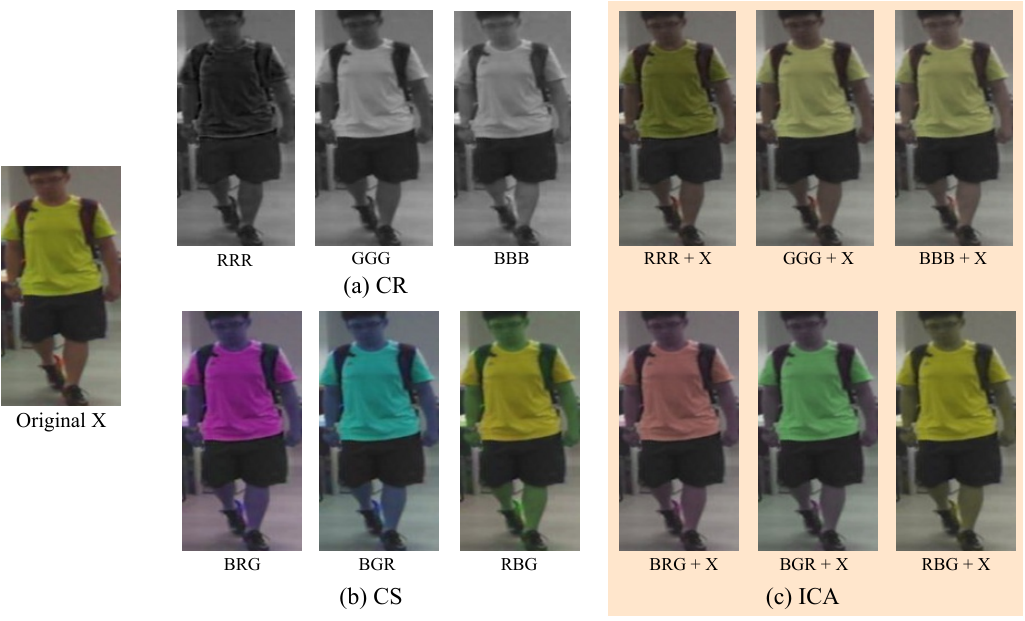}
  \vspace{-1mm}
  \caption{Illustration of the Channel Replacement Augmentation (CR), our Channel Swap Augmentation (CS), and final Image-Level Color-Augmentation (ICA). Best viewed in color.}
  \label{fig:channel}
  \vspace{-3mm}
\end{figure}

However, CR transfers the RGB images into IR-like images, which only focuses on the VI-ReID task (refer to Fig.~\ref{fig:channel}(a)). It introduces limited variation in the color and is unsuitable for CC-ReID. Consequently, we propose Channel Swap Augmentation (CS) to randomly change the order of three color channels, which generates more color variation of the same identity. This process keeps all the information of input RGB images while increasing the appearance variation. The CS is formulated as:
\begin{equation}
    \begin{aligned}
    \mathbf{\tilde{x}}_{i}^{V} =\{x^{C_1},x_{i}^{C_2},x_{i}^{C_3}\},\ \{C_1,C_2,C_3\} \in \textit{\textbf{S}}\{R,G,B\},
    \end{aligned}
\label{eq:ica2}
\end{equation}
where \textit{\textbf{S}} denotes the set of shuffling orders of R, G, and B.

After the CS operation, we can see drastic changes in the image color (refer to Fig.~\ref{fig:channel}(b)). With the image color constantly changing for the same identity, the model learns more color-insensitive features. However, the CS-generated images might change the unwanted region color of identities (\textit{e.g.} exposed skin). To solve this aforementioned problem, we follow the Mix-Up method~\cite{zhang2017mixup} and directly fuse the augmented image with its original one. Our Image-Level Color-Augmentation (ICA) can be defined as:
\begin{equation}
  \mathbf{\tilde{x}}_{i}^{V'}=0.5 \times \mathbf{x}_{i}^{V} + 0.5 \times \mathbf{\tilde{x}}_{i}^{V},
\label{eq:ica3}
\end{equation}
all augmented images are more realistic after ICA, as shown in Fig.~\ref{fig:channel}(c).

For VI-ReID, the input will be the original RGB images with its ICA-augmented pair and grouped up with the corresponding IR images. The learning objective of VI-ReID is defined as:
\begin{equation}
  \sum \mathcal{L}(f^{V}(\mathbf{x}_{i}^{V}),f^{V}(\mathbf{\tilde{x}}_{i}^{V'}),f^{I}(\mathbf{x}_{j}^{I}),\;y_{i},y_{i},y_{j}),
\vspace{-2mm}
\end{equation}
where $y_{i}$ and $y_{j}$ are the labels for RGB and IR images, respectively. $\mathcal{L}(\cdot)$ is the Loss function for optimizing. $f^{V}(\cdot)$ is the learning architecture in RGB stream, and $f^{I}(\cdot)$ is that in IR stream. As our ICA augmentation performs very similarly to the cloth-changing scenario visually, we also take it into CC-ReID. The learning objective of CC-ReID is:

\begin{equation}
  \sum \mathcal{L}(f^{V}(\mathbf{x}_{i}^{V}),f^{V}(\mathbf{\tilde{x}}_{i}^{V'}),\;y_{i},y_{i}).
\vspace{-2mm}
\end{equation}

\subsection{Pixel-Level Color-Transformation} 
\begin{figure}[ht]
  \centering
  \includegraphics[width=0.9\linewidth]{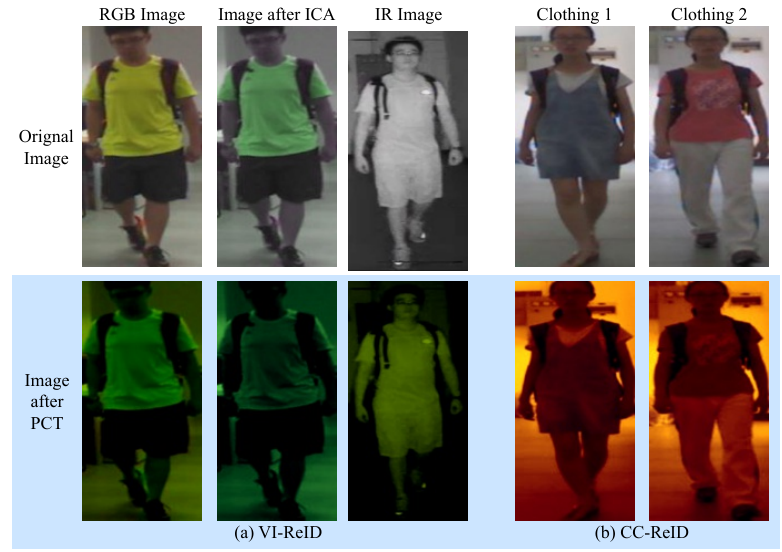}
  \vspace{-2mm}
  \caption{Visualization of input images after PCT. (a) VI-ReID: Original images in RGB and IR modalities and corresponding ones after PCT. (b) CC-ReID: Original images with different clothing colors and corresponding ones after PCT. Regardless of color difference, PCT manages to transform the input into a new common color space. Best viewed in color.}
  \vspace{-2mm}
  \label{fig:pct}
\end{figure}

\noindent To further reduce the color variation influence, we design Pixel-Level Color-Transformation (PCT) to project the input images onto a new common color space. Specifically, we adopt a non-linear activation function and Batch Normalization (BN) in PCT to make the learning strategy more effective. For VI-ReID, PCT is designed in a two-stream structure to fuse the cross-modality features. We design a separated intra-modality learning stage in the first $Conv$ layer and the second cross-modality learning $Conv$ with the sharing weights, which is represented by:
\begin{gather}
\begin{aligned}
    \mathbf{\tilde{\tilde{x}}}_{i}^{V}, \mathbf{\tilde{\tilde{x}}}_{i}^{V'}&={C}^{x}(\sigma(BN({C}^{in}(\mathbf{x}_{i}^{V},\mathbf{\tilde{x}}_{i}^{V'})))),\\
    \mathbf{\tilde{\tilde{x}}}_{i}^{I}&={C}^{x}(\sigma(BN({C}^{in}(\mathbf{x}_{i}^{I})))),
\end{aligned}
\label{eq:pct-vi-reid}
\end{gather}
where ${C}^{in}$ represents the first intra-modality learning stage, and ${C}^{x}$ represents the second cross-modality learning stage. Every $C$ is a $1\times1$ $Conv$ layer. Besides, $\sigma$ represents the nonlinear $ReLU$ activation function. After PCT, the color-transformed images are adaptive to any deep learning backbone. Results of images after PCT are shown in Fig.~\ref{fig:pct}. The cross-modality or cloth-changing images after PCT look much more similar. It significantly reduces the color discrepancies and also demonstrates that color-irrelevant information is extracted by our module. It is worth noting that the IR-stream is only adopted in VI-ReID.

\subsection{Backbone and Loss Function}
\label{ssec:loss}
\noindent ResNet50 backbone~\cite{he2016deep} is used in our CSL method for extracting features from input images. We assume that the color-inconsistent discrepancies mainly lie in low-level features~\cite{ye2021deep}, we use separate parameters for shallow layers while sharing the remaining ones. For CC-ReID, only RGB stream is adopted. This architecture beneﬁts from a uniform structure that simultaneously captures the cross-modality information and generates common features to solve two tasks together. Inspired by the success of previous works~\cite{ye2021deep,ye2021channel}, we also incorporated the Non-Local~\cite{xia2019second} into our backbone.

For the loss function, we adopt identity classification loss ($\mathcal{L}_{id}$), which is widely used in~\cite{ye2018hierarchical,ye2021channel,ye2021deep}:
\vspace{-2mm}
\begin{equation}
    \mathcal{L}_{id}=-\frac{1}{N}\sum_{i=1}^{N}log((y_i\,|\,f^{M}(\mathbf{x}_{i}^{M})))\quad M\in\{R,I\},
\label{eq:idloss}
\end{equation}
where $f^{M}(\mathbf{x}_{i}^{M})$ is the extracted features of input images, and $y_i$ is the corresponding labels.

We also bring in enhanced squared difference loss ($\mathcal{L}_{sq}$)~\cite{ye2021channel}, which is developed based on weighted regularization triplet loss ($\mathcal{L}_{wrt}$)~\cite{ye2021deep}:
\begin{equation}
\begin{aligned}
    &\mathcal{L}_{sq} = -\frac{1}{N}\sum_{i=1}^{N}log(1+exp[\phi(\delta)]),\\
    &\delta = \sum_{ij}^{}w_{ij}^{p}d_{ij}^{p}-\sum_{ij}^{}w_{ik}^{n}d_{ik}^{n},\ \ \ 
    \phi[\delta] =
    \left\{\begin{array}{l}
      \delta^2\quad \text{if}\;\delta>0 \\
      -\delta^2\quad \text{if}\;\delta<0,
    \end{array}\right.\\
    &w_{ij}^{p} = \frac{exp(d_{ij}^{p})}{\sum exp(d_{ij}^{p})},\quad w_{ik}^{n} = \frac{exp(d_{ik}^{n})}{\sum exp(d_{ik}^{n})},
\end{aligned}
\label{eq:sqloss}
\end{equation}
where $(i,j,k)$ represents a triplet within each training batch. For anchor $\mathbf{x}_{i}$, $p$ represents the positive set and $n$ is the negative set. $d_{ij} = \left\|f^{M}(\mathbf{x}_{i}^{M})-f^{M}(\mathbf{x}_{j}^{M})\right\|_{2}$ is the Euclidean distance between two samples. $w_{ij}^{p}$ and $w_{ik}^{n}$ are the regularized weights for positive and negative pairs respectively. Compared with traditional triplet loss, $\mathcal{L}_{sq}$ increases the contribution of the hard triplets ($\delta\in[-1,0]$) and reduces the effect of easy triplets ($\delta<-1$). Inspired by~\cite{ye2021channel}, we combine $\mathcal{L}_{id}$ and $\mathcal{L}_{sq}$ together as our overall training objective, denoted by:
\begin{equation}
    \mathcal{L}_{total}=\mathcal{L}_{id}+\mathcal{L}_{sq}.
\label{eq:totalloss}
\end{equation}

\section{NTU-Corridor Benchmark}
\label{sec:dataset}
\noindent As shown in Tab.~\ref{tab:vi-reid-datasets}, existing VI-ReID datasets (\textit{i.e.} \textit{SYSU-MM01}~\cite{wu2017rgb} and  \textit{RegDB}~\cite{nguyen2017person}) only involve a limited number of cameras and environments, and all RGB images are captured under sufficient illuminance. However, the illuminance variation is a challenging problem in ReID tasks~\cite{zhang2022illumination}. Besides, existing datasets only contain ground-view images, but a top-down view is closer to real applications. RegDB~\cite{nguyen2017person} uses thermal sensors, but thermal images usually contain limited texture or pattern information of the objects, as shown in Fig.~\ref{fig:dataset}(c). Conversely, most existing surveillance cameras use NIR cameras during night-time to record more subject information. As shown in Fig.~\ref{fig:dataset}(b), NIR images in SYSU-MM01~\cite{wu2017rgb} keep more information (\textit{e.g.} clothing patterns).

To solve the aforementioned problems, we construct a new NIR VI-ReID dataset: \textbf{NTU-Corridor}. We capture images from surveillance cameras in NTU buildings. Those cameras can automatically switch to NIR mode when illuminance is insufficient. Our dataset consists of both indoor and outdoor scenes with large changes in viewpoint, illumination, and resolution that manifest even within individual wide-angle cameras. Specifically, it spans 136 cameras, significantly more than other benchmarks. There are 128 appearance identities with signed privacy agreements for using their images for academic purposes. To the best of our knowledge, it is the only public VI-ReID dataset with privacy agreements. The statistical information of our dataset is shown in Tab.~\ref{tab:vi-reid-datasets}. Some sample images are available in Fig.~\ref{fig:dataset}(a).

\begin{table}[h]
  \centering
  \vspace{-1mm}
  \caption{The statistics of the NTU-Corridor and other public VI-ReID datasets. More information is in appendix.}
  \vspace{-1mm}
  \scalebox{0.86}{\begin{tabular}{@{}c||c||c||c@{}}
    \toprule
    \multirow{3}*{Datasets} & \multirow{3}*{\shortstack{NTU-\\Corridor}} & \multirow{3}*{\shortstack{SYSU-\\MM01\\\cite{wu2017rgb}}} & \multirow{3}*{\shortstack{RegDB\\\cite{nguyen2017person}}}\\
    ~ & ~ & ~ & ~\\
    ~ & ~ & ~ & ~\\\cline{1-4}
    IR Type & \textbf{NIR} & \textbf{NIR} & Thermal\\
    View & \textbf{Top-Down} & Ground & Ground\\
    \# Identity & 128 & \textbf{491} & 412\\
    \# Image & 29783 & \textbf{303420} & 8420\\
    \# Camera & \textbf{136} & 6 & 2\\
    \# Environment & \textbf{48} & 5 & 1\\
    Illumination & \textbf{Variable} & Good & Good\\
    Image Size & \textbf{Vary} & \textbf{Vary} & Fix\\
    Avg. Size & \textbf{184×456} & 111×291 & 64×128\\
    \bottomrule
  \end{tabular}}
  \label{tab:vi-reid-datasets}
  \vspace{-2mm}
\end{table}

\section{Experiments}
\label{sec:exp}
\subsection{Datasets and Evaluation Protocols}

\noindent\textbf{Visible-Infrared Person ReID (VI-ReID):} We evaluate our CSL on three VI-ReID datasets: SYSU-MM01~\cite{wu2017rgb}, RegDB~\cite{nguyen2017person}, and our NTU-Corridor. The statistical information of them is listed in Tab.~\ref{tab:vi-reid-datasets}. The official testing phase of SYSU-MM01 focuses on the performance of the NIR-to-RGB setting (NIR images are used as query, and RGB images are gallery sets). However, in real-world surveillance systems, bi-directional searching is a common requirement. With this insight, we propose an additional RGB-to-NIR evaluation protocol for SYSU-MM01 and NTU-Corridor (RGB images are query set, and NIR images are gallery set).

\noindent \textbf{Cloth-Changing Person ReID (CC-ReID):} We also evaluate our method on two widely-used CC-ReID datasets:  PRCC~\cite{yang2019person} and LTCC~\cite{qian2020long}. PRCC~\cite{yang2019person} is captured under 3 camera views and identities in different clothing sets. There are 33698 images, with 150 identities in the training set and 71 identities in the testing set. LTCC~\cite{qian2020long} contains 17138 images of 152 identities. The training set has 77 identities, including 46 in different clothing sets, while the testing set consists of 45 clothing-change identities.

\noindent\textbf{Evaluation Metrics:} We adopt the most common Rank 1 Accuracy (Rank1) and mean Average Precision (mAP) as the evaluation metrics for all our experiments.

\subsection{Implementation Details}
\label{ssec:implement}
\noindent We implemented our model on the PyTorch framework. Following the structure of~\cite{ye2021deep}, we adopted a Non-Local~\cite{xia2019second} enhanced ResNet50~\cite{he2016deep} as the backbone. The initial model is pre-trained on ImageNet~\cite{deng2009imagenet}. We apply random cropping for all the training images. For SYSU-MM01~\cite{wu2017rgb} and NTU-Corridor, we follow the same setting as \cite{ye2021channel} to resize images into $288\times144$. The original image sizes in RegDB~\cite{nguyen2017person} are fixed in $128\times64$, thus we resize them in a smaller size, $208\times104$. As for CC-ReID, we resize the images to $384\times192$, following the standard protocol~\cite{qian2020long}. We train our model for 100 epochs with a batch size of 32, using the SGD with a 0.9 momentum and a 5e-4 weight decay. The initial learning rate is 0.1, and decayed by 0.1 and 0.01 at 20 and 50 epochs. We also apply a warm-up strategy~\cite{luo2019strong} in the first 10 epochs. In each batch, we randomly sample 4 identities with 4 images in each modality.

\subsection{Comparison with VI-ReID State-of-the-Art}
\label{ssec:vi-sota}
\noindent\textbf{SYSU-MM01 Dataset:} We first compare our CSL method with VI-ReID state-of-the-art methods on the SYSU-MM01 dataset. Following the official evaluation metrics, performances of our CSL and state-of-the-art methods are compared in Tab.~\ref{tab:vi-reid-results-sysu1}~\footnote{More discussion about RGB-to-NIR setting is in appendix.}. The results of CSL outperform MAUM~\cite{liu2022learning} 0.8\% and 3.2\% on Rank1 of \textit{All-search} and \textit{Indoor-Search} settings, respectively.

\begin{table}[t]
  \centering
  \caption{State-of-the-art Comparison for VI-ReID on SYSU-MM01~\cite{wu2017rgb} on conventional NIR-to-RGB evaluation. Only some methods are listed here, the entire table is listed in appendix.}
  \vspace{-2mm}
    \scalebox{0.86}{\begin{tabular}{@{}c||cc|cc@{}}
    \toprule
    \multirow{3}*{Methods} & \multicolumn{4}{c}{SYSU-MM01 \textit{(NIR to RGB)}}\\\cline{2-5}
    ~ & \multicolumn{2}{c|}{All-search} & \multicolumn{2}{c}{Indoor-Search}\\\cline{2-5}
    ~ & Rank1 & mAP & Rank1 & mAP\\ \cline{1-5}
    AGW\cite{ye2021deep} & 47.5 & 47.7 & 54.2 & 63.0\\
    XIV\cite{li2020infrared} & 49.9 & 50.7 & - & -\\
    LbA\cite{park2021learning} & 55.4 & 54.1 & 58.5 & 66.3\\
    CAJ\cite{ye2021channel} & 69.9 & 66.9 & 76.3 & 80.4\\
    MAUM\cite{liu2022learning} & 71.7 & \textbf{68.8} & 77.0 & 81.9\\
    \textbf{CSL} & \textbf{72.5} & 68.0 & \textbf{80.2} & \textbf{82.9}\\
    \bottomrule
  \end{tabular}}
  \label{tab:vi-reid-results-sysu1}
  \end{table}

\noindent \textbf{NTU-Corridor Dataset:} Then we compare our CSL methods with state-of-the-art methods on NTU-Corridor dataset, and the results are shown in Tab.~\ref{tab:vi-reid-results-ntu}. We also conduct experiments on the bi-direction settings to mimic the real-world scenarios. Our CSL methods outperform previous methods consistently on all the evaluation metrics.

\noindent \textbf{RegDB Dataset:} It is a widely used Thermal dataset for VI-ReID, and the experiment results on RegDB are shown in \autoref{tab:vi-reid-results-regdb}. For thermal images, most texture information is not present, and thus NIR sensors are used in most surveillance cameras nowadays instead of Thermal ones. Because our CSL approach only removes the color information during the training, it focuses more on the texture, contour, and body-shape features, which is more suitable for NIR VI-ReID. Nevertheless, our CSL still manages to get the second-best result compared with other state-of-the-art methods in RegDB.

\begin{table}[h]
  \centering
  \vspace{-1mm}
  \caption{State-of-the-art Comparison for VI-ReID on RegDB~\cite{nguyen2017person}. Only some methods are listed here, the entire table is listed in appendix.}
  \vspace{-1mm}
  \scalebox{0.86}{\begin{tabular}{@{}c||cc|cc@{}}
    \toprule
    \multirow{3}*{Methods} & \multicolumn{4}{c}{RegDB} \\  \cline{2-5}
    ~ & \multicolumn{2}{c|}{\textit{Thermal-to-RGB}} & \multicolumn{2}{c}{\textit{RGB-to-Thermal}}\\ \cline{2-5}
    ~ & Rank1 & mAP & Rank1 & mAP\\ \cline{1-5}
    JSIA\cite{wang2020cross} & 48.1 & 48.9 & 48.5 & 49.3\\
    AlignGAN\cite{wang2019rgb} & 56.3 & 53.4 & 57.9 & 53.6\\
    XIV\cite{li2020infrared} & - & - & 62.2 & 60.2\\
    AGW\cite{ye2021deep} & - & - & 70.1 & 66.4\\
    LbA\cite{park2021learning} & 72.4 & 65.5 & 74.2 & 67.6\\
    CAJ\cite{ye2021channel} & 84.8 & 77.8 & 85.0 & 79.1\\
    MAUM\cite{liu2022learning} & \textbf{87.0} & \textbf{84.3} & \textbf{87.9} & \textbf{85.1}\\
    \textbf{CSL} & \underline{85.8} & 77.8 & \underline{86.2} & 77.9\\
    \bottomrule
  \end{tabular}}
  \vspace{-2mm}
  \label{tab:vi-reid-results-regdb}
\end{table}

\begin{table}[h]
    \centering
    \vspace{-1mm}
    \caption{State-of-the-arts Comparison for VI-ReID on NTU-Corridor.}
    \vspace{-1mm}
    \scalebox{0.86}{\begin{tabular}{@{}c||cc|cc@{}}
    \toprule
    \multirow{3}*{Methods} & \multicolumn{4}{c}{NTU-Corridor}\\\cline{2-5}
    ~ & \multicolumn{2}{c|}{\textit{NIR-to-RGB}} & \multicolumn{2}{c}{\textit{RGB-to-NIR}}\\\cline{2-5}
    ~ & Rank1 & mAP & Rank1 & mAP\\ \cline{1-5}
    LbA\cite{park2021learning} & 64.3 & 47.7 & 68.4 & 45.9\\
    AGW\cite{ye2021deep} & 68.2 & 47.0 & 68.2 & 41.2\\
    CAJ\cite{ye2021channel} & 82.0 & 63.3 & 83.3 & 60.5\\
    \textbf{CSL} & \textbf{83.4} & \textbf{64.9} & \textbf{86.2} & \textbf{63.7}\\
    \bottomrule
    \end{tabular}}
    \label{tab:vi-reid-results-ntu}
    \vspace{-2mm}
\end{table}

\subsection{Comparison with CC-ReID State-of-the-Art}
\label{ssec:cc-sota}
\noindent As shown in Tab.~\ref{tab:cc-reid-results}, our CSL method outperforms state-of-the-art methods in LTCC and PRCC by a large margin. Both VI-ReID and CC-ReID results prove the learned color-invariant representation from our method is robust against cross-modality and cloth-changing scenarios.
\begin{table}[h]
  \centering
  \caption{State-of-the-art Comparison for CC-ReID on PRCC~\cite{yang2019person} and LTCC~\cite{qian2020long}. Only some methods are listed here, the entire table is listed in appendix.}
  \vspace{-1mm}
  \scalebox{0.86}{\begin{tabular}{@{}c||cc||cc@{}}
    \toprule
    \multirow{2}*{Methods} & \multicolumn{2}{c||}{LTCC} & \multicolumn{2}{c}{PRCC} \\ \cline{2-5}
    ~ & Rank1 & mAP & Rank1 & mAP\\ \cline{1-5}
    3DSL\cite{chen2021learning} & 31.2 & 14.8 & - & 51.3\\
    FSAM\cite{hong2021fine} & 38.5 & 16.2 & 54.5 & -\\
    CAL\cite{gu2022clothes} & 40.1 & 18.0 & 55.2 & 55.8\\
    \textbf{CSL} & \textbf{56.2} & \textbf{22.2} & \textbf{56.4} & \textbf{56.0}\\
    \bottomrule
  \end{tabular}}
  \label{tab:cc-reid-results}
  \vspace{-3mm}
\end{table}

\subsection{Ablation Study}
\label{ssec:ablation}
\noindent In this section, we evaluate the performance of each component of CSL on both VI-ReID and CC-ReID benchmarks in Tab.~\ref{tab:ablation-study}. The AGW framework with the proposed enhanced squared difference loss ($\mathcal{L}_{sq}$) is adopted as our baseline. Rank1 accuracy of the baseline with our ICA is increased by 6.2\% and 7.2\% on All-Search and Indoor-Search settings of SYSU-MM01. It also increases 4.6\% and 1.1\% on LTCC~\footnote{We also compare our ICA with other data-augmentation strategies and discuss the results in appendix.}. To discover the relation between the ICA and PCT, we conduct experiments with baseline and PCT. However, only introducing PCT leads to worse results than ICA, and the results are even worse than the baseline for LTCC. It is easy to understand that PCT can project images into a new color space only when provided with enough color variant samples. Consequently, ICA is an essential module for PCT to learn color-insensitive features. When we perform ICA and PCT simultaneously, the Rank1 accuracy is further improved by 3.8\% and 6.0\% on SYSU-MM01. Besides, it also increases by 9.3\% on LTCC. Therefore, PCT can help to shrink the cross-modality and cloth-changing discrepancies when combined with ICA~\footnote{More discussion about ICA and PCT is in appendix.}. Ultimately, our method achieves the highest performances on VI-ReID and CC-ReID tasks simultaneously.

\begin{table}[h]
  \centering
  \caption{The influence of each component of our proposed CSL approach. Experiments are conducted on both VI-ReID and CC-ReID wide-used benchmarks. Results on NTU-Corridor and PRCC are listed in appendix.}
  \vspace{-1mm}
  \scalebox{0.86}{\begin{tabular}{@{}l||cc|cc||cc@{}}
    \toprule
     \multirow{3}*{Experiment Setting} & \multicolumn{4}{c||}{SYSU-MM01} & \multicolumn{2}{c}{LTCC}\\ \cline{2-7}
     ~ & \multicolumn{2}{c|}{All-Search} & \multicolumn{2}{c||}{Indoor-Search} & \multirow{2}*{Rank1} & \multirow{2}*{mAP}\\ \cline{2-5}
     ~ & Rank1 & mAP & Rank1 & mAP\\ \cline{1-7}
     Baseline & 62.5 & 59.1 & 67.0 & 72.4 & 42.3 & 15.9\\
     Baseline+ICA & 68.7 & 64.7 & 74.2 & 78.5 & 46.9 & 16.6\\
     Baseline+PCT & 65.8 & 62.5 & 70.7 & 74.8 & 37.5 & 16.0\\
     Baseline+ICA+PCT & \textbf{72.5} & \textbf{68.0} & \textbf{80.2} & \textbf{82.9} & \textbf{56.2} & \textbf{22.2}\\
     \bottomrule
  \end{tabular}}
  \label{tab:ablation-study}
  \vspace{-2mm}
\end{table}

\section{Conclusion}
\label{sec:con}

This paper proposes Color Space Learning (CSL) for Cross-Color Person Re-identification tasks. We guide the network to focus on color-insensitive features with our proposed Image-Level Color-Augmentation and Pixel-Level Color-Transformation module. We also contribute a new Campus-Corridor benchmark with privacy agreements, which is captured by 136 ceiling-mounted surveillance cameras. Extensive experiments demonstrate the effectiveness and superiority of our approach in color-insistent circumstances due to cross-modality and cloth-changing.

\section*{Acknowledgment}

This research work was carried out at the Rapid-Rich Object Search (ROSE) Lab, Nanyang Technological University, Singapore. The research is partially supported by the Defence Science and Technology Agency (DSTA), under the project agreement No. DST000ECI22000195.

\section*{Appendix for ``Color Space Learning for Cross-Color Person Re-Identification"}

\setcounter{table}{0}
\setcounter{figure}{0}
\setcounter{section}{0}
\setcounter{equation}{0}
\renewcommand{\thetable}{A\arabic{table}}
\renewcommand{\thefigure}{A\arabic{figure}}
\renewcommand{\thesection}{A\arabic{section}}
\renewcommand{\theequation}{A\arabic{equation}}

\section{Campus-Corridor Benchmark}

We discuss more details of our constructed Campus-Corridor benchmark in this section. 

\subsection{Evaluation Protocol}

\noindent We divide the Campus-Corridor dataset into a training set and a test set at a ratio around 1.3:1. The training set contains 17,006 images of 74 identities (14,695 images are from the RGB modality and 2,311 are from the IR modality), and the test set contains 12,777 images of 54 identities (11,143 images are from the RGB modality and 1,634 are from the IR modality). To make our benchmark imitate real-world ReID circumstances, both the RGB-to-NIR and the NIR-to-RGB scenarios are used to evaluate the performance of the VI-ReID methods. We inherit the most common Rank 1 Accuracy (Rank1) and mean Average Precision (mAP) as the evaluation metrics. During the test stage, we randomly choose one image from the images of each identity to form the gallery set for evaluating the performance of the models.

\subsection{Statistical Distribution}

\noindent For a ReID dataset, the statistical distribution is very important. Consequently, we briefly discuss our Campus-Corridor dataset in two statistical aspects, the appearance ratio and the gender ratio.

\begin{figure}[h]
\centering
\includegraphics[width=1\linewidth]{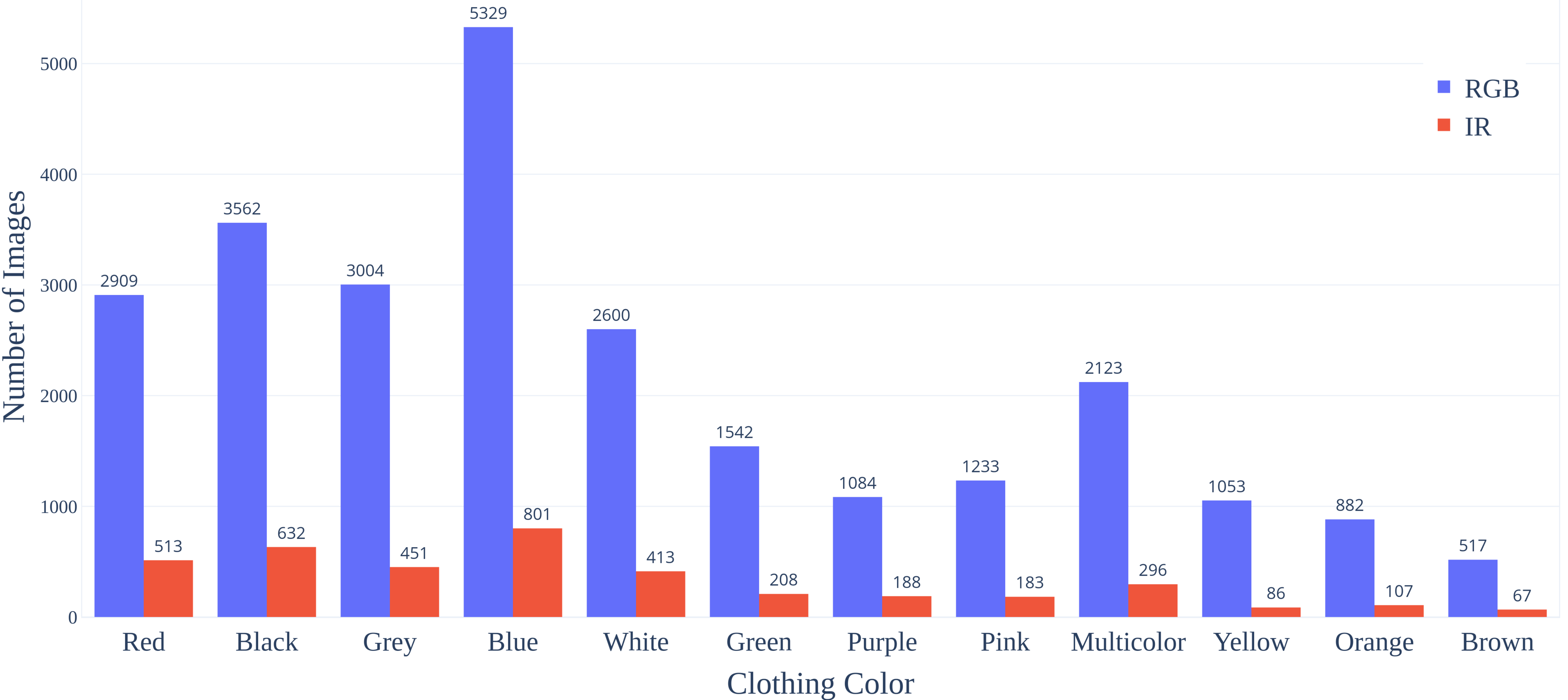}
\vspace{-4mm}
\caption{Clothing color distribution in the Campus-Corridor dataset. A handful of common colors such as blue and black dominate the population, leading to a clothing color skew.}
\label{fig:color}
\vspace{-2mm}
\end{figure}

The appearance attribute labels of persons are extremely valuable auxiliary information. Integrating attribute information during the model training stage can yield more generalized and robust feature representations of different persons \cite{yang2019person, qian2020long}. Most of the Person ReID datasets did not come with attribute labels initially. Some attribute annotations (such as Market-1501~\cite{zheng2015scalable} and DukeMTMC-reID~\cite{zheng2017unlabeled} attributes labels) are completed by third-party organizations. The attribute annotation procedure is tedious, expensive, and time-consuming. In our Campus-Corridor dataset, most of the appearance and soft-biometric attributes are labeled by the participants themselves. This eliminates the manual attribute annotation process. 

The color distribution of the cloth of the pedestrians is demonstrated in Fig.~\ref{fig:color} below. Black, blue, and grey are the dominant colors for the upper body. These three colors combined contribute to more than 50\% of the total upper body color labels. In addition, our dataset also provides other attributes like clothing types.

\begin{figure}[h]
\centering
\includegraphics[width=0.85\linewidth]{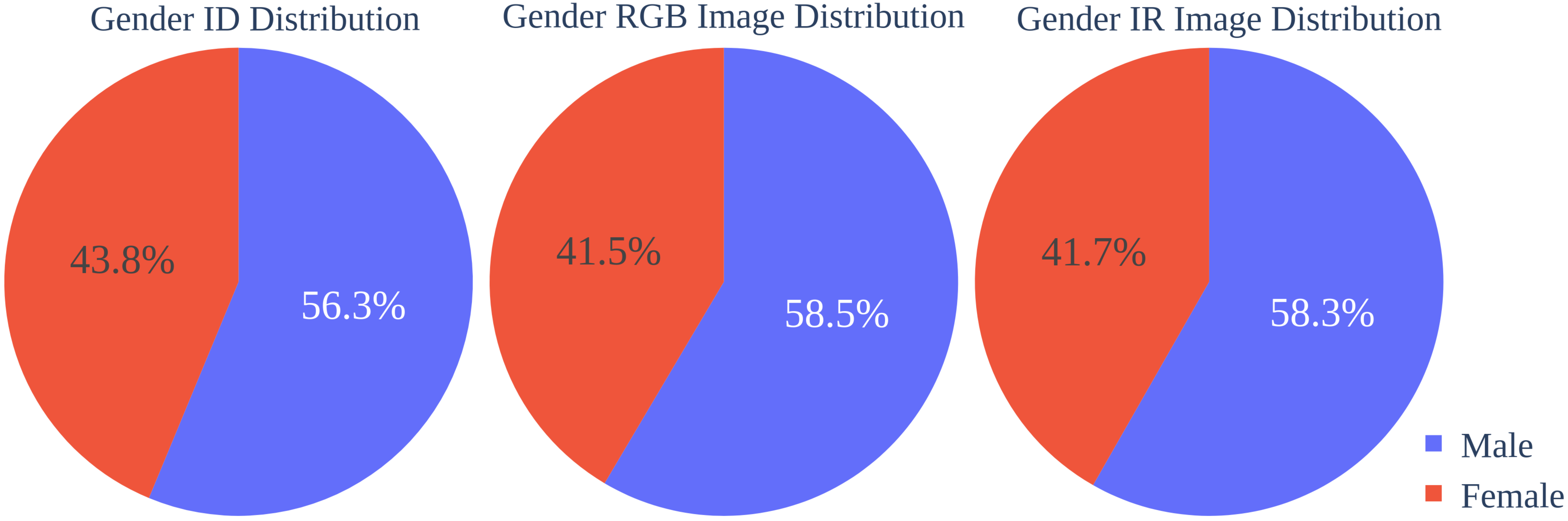}
\caption{Gender distribution in Campus-Corridor dataset. The population captured in our cameras is a gender-balanced environment and is reflected in our dataset's distribution.}
\label{fig:gender}
\vspace{-2mm}
\end{figure}

To prevent gender bias in the Person ReID models, we try to control the male and female ratio during the data collection. In the Campus-Corridor dataset, the male and female ratio of the participants and the images are very well balanced, shown in Fig.~\ref{fig:gender}.

\section{Experiments}

In this section, we list the entire experimental results and make more discussion.

\subsection{Comparasion with VI-ReID State-of-the-Art}

\noindent\textbf{SYSU-MM01 Dataset:} Following the official evaluation metrics, performances of our CSL and all state-of-the-art methods are compared in Tab.~\ref{tab:vi-reid-results-sysu1}. The results of CSL outperform all previous methods consistently. Moreover, to systematically compare and analyze the methods, we conduct experiments on our newly proposed RGB-to-NIR setting, and the results are shown in Tab.~\ref{tab:vi-reid-results-sysu2}. Our CSL coherently outperforms other methods by a margin. We can find that our approach achieves very similar results on both NIR-to-RGB and RGB-to-NIR settings on SYSU-MM01. The challenge of these settings is that the RGB images are much more than NIR ones in the SYSU-MM01 dataset, \textit{i.e.}, the data-imbalance problem. The results prove that our method is adaptive for challenging circumstances compared with GAN-based methods (\textit{i.e.}, AlignGAN~\cite{wang2019rgb} and JSIA~\cite{wang2020cross}), the results of which are different in bi-directional searching.

\begin{table}[t]
  \centering
  \caption{State-of-the-art Comparison for VI-ReID on SYSU-MM01~\cite{wu2017rgb} on conventional NIR-to-RGB evaluation.}
  \vspace{-1mm}
    \scalebox{0.85}{\begin{tabular}{@{}c||cc|cc@{}}
    \toprule
    \multirow{3}*{Methods} & \multicolumn{4}{c}{SYSU-MM01 \textit{(NIR to RGB)}}\\\cline{2-5}
    ~ & \multicolumn{2}{c|}{All-search} & \multicolumn{2}{c}{Indoor-Search}\\\cline{2-5}
    ~ & Rank1 & mAP & Rank1 & mAP\\ \cline{1-5}
    HCML\cite{ye2018hierarchical} & 14.3 & 16.2 & 24.5 & 30.1\\
    Zero-Padd.\cite{wu2017rgb} & 14.8 & 15.9 & 20.6 & 26.9\\
    BDTR\cite{ye2018visible} & 17.0 & 19.7 & - & -\\
    cm-GAN\cite{dai2018cross} & 26.9 & 27.8 & 31.6 & 42.2\\
    JSIA\cite{wang2020cross} & 38.1 & 36.9 & 43.8 & 52.9\\
    AlignGAN\cite{wang2019rgb} & 42.4 & 40.7 & 45.9 & 54.3\\
    AGW\cite{ye2021deep} & 47.5 & 47.7 & 54.2 & 63.0\\
    XIV\cite{li2020infrared} & 49.9 & 50.7 & - & -\\
    DDAG\cite{ye2020dynamic} & 54.8 & 53.0 & 61.0 & 68.0\\
    LbA\cite{park2021learning} & 55.4 & 54.1 & 58.5 & 66.3\\
    NFS\cite{chen2021neural} & 56.9 & 55.5 & 62.8 & 69.8\\
    CICL\cite{zhao2021joint} & 57.2 & 59.3 & 66.6 & 74.7\\
    FMI\cite{tian2021farewell} & 60.0 & 58.8 & 66.1 & 73.0\\
    cm-SSFT\cite{lu2020cross} & 61.6 & 63.2 & 70.5 & 72.6\\
    CM-NAS\cite{fu2021cm} & 62.0 & 60.0 & 67.0 & 73.0\\
    SPOT\cite{chen2022structure} & 65.3 & 62.3 & 69.4 & 74.6\\
    MCLNet\cite{hao2021cross} & 65.4 & 62.0 & 72.6 & 76.6\\
    FMCNet\cite{zhang2022fmcnet} & 66.3 & 62.5 & 68.2 & 74.1\\
    SMCL\cite{wei2021syncretic} & 67.4 & 61.8 & 68.8 & 75.6\\
    DART\cite{yang2022learning} & 68.7 & 66.2 & 72.5 & 78.2\\
    CAJ\cite{ye2021channel} & 69.9 & 66.9 & 76.3 & 80.4\\
    MAPNet\cite{wu2021discover} & 70.6 & 68.2 & 76.7 & 81.0\\
    MAUM\cite{liu2022learning} & 71.7 & \textbf{68.8} & 77.0 & 81.9\\
    \textbf{CSL} & \textbf{72.5} & 68.0 & \textbf{80.2} & \textbf{82.9}\\
    \bottomrule
  \end{tabular}}
  \label{tab:vi-reid-results-sysu1}
  \end{table}

\vspace{1em}
\begin{table}[t]
    \centering
    \caption{State-of-the-art Comparison for VI-ReID on SYSU-MM01~\cite{wu2017rgb} on our newly proposed RGB-to-NIR evaluation.}
    \vspace{-2mm}
    \scalebox{0.85}{\begin{tabular}{@{}c||cc|cc@{}}
    \toprule
    \multirow{3}*{Methods} & \multicolumn{4}{c}{SYSU-MM01 \textit{(RGB-to-NIR)}}\\\cline{2-5}
    ~ & \multicolumn{2}{c|}{All-search} & \multicolumn{2}{c}{Indoor-Search}\\\cline{2-5}
    ~ & Rank1 & mAP & Rank1 & mAP\\ \cline{1-5}
    JSIA\cite{wang2020cross} & 31.2 & 40.6 & 39.5 & 54.3\\
    AlignGAN\cite{wang2019rgb} & 35.0 & 45.2 & 44.7 & 60.2\\
    AGW\cite{ye2021deep} & 45.9 & 52.1 & 48.7 & 52.6\\
    LbA\cite{park2021learning} & 56.2 & 58.6 & 57.8 & 59.6\\
    CAJ\cite{ye2021channel} & 69.5 & 69.3 & 72.6 & 71.8\\
    \textbf{CSL} & \textbf{73.5} & \textbf{72.4} & \textbf{78.5} & \textbf{76.9}\\
    \bottomrule
  \end{tabular}}
  \label{tab:vi-reid-results-sysu2}
\end{table}

\noindent \textbf{RegDB Dataset:} We compare our CSL with all previous methods on RegDB, and the performances are shown in \autoref{tab:vi-reid-results-regdb}.

\begin{table}[h]
  \centering
  \caption{State-of-the-art Comparison for VI-ReID on RegDB~\cite{nguyen2017person}.}
  \vspace{-2mm}
  \scalebox{0.85}{\begin{tabular}{@{}c||cc|cc@{}}
    \toprule
    \multirow{3}*{Methods} & \multicolumn{4}{c}{RegDB} \\  \cline{2-5}
    ~ & \multicolumn{2}{c|}{\textit{Thermal-to-RGB}} & \multicolumn{2}{c}{\textit{RGB-to-Thermal}}\\ \cline{2-5}
    ~ & Rank1 & mAP & Rank1 & mAP\\ \cline{1-5}
    Zero-Padd.\cite{wu2017rgb} & 16.6 & 17.8 & 17.8 & 18.9\\
    HCML\cite{ye2018hierarchical} & 21.7 & 22.2 & 24.4 & 20.1\\
    BDTR\cite{ye2018visible} & 32.7 & 31.1 & 33.5 & 31.8\\
    JSIA\cite{wang2020cross} & 48.1 & 48.9 & 48.5 & 49.3\\
    AlignGAN\cite{wang2019rgb} & 56.3 & 53.4 & 57.9 & 53.6\\
    XIV\cite{li2020infrared} & - & - & 62.2 & 60.2\\
    DDAG\cite{ye2020dynamic} & 68.0 & 61.8 & 69.3 & 63.5\\
    AGW\cite{ye2021deep} & - & - & 70.1 & 66.4\\
    cm-SSFT\cite{lu2020cross} & 71.0 & 71.7 & 72.3 & 72.9\\
    FMI\cite{tian2021farewell} & 71.8 & 70.1 & 73.2 & 71.6\\
    LbA\cite{park2021learning} & 72.4 & 65.5 & 74.2 & 
67.6\\
    MCLNet\cite{hao2021cross} & 75.9 & 69.5 & 80.3 & 73.1\\
    CICL\cite{zhao2021joint} & 77.9 & 69.4 & 78.8 & 69.4\\
    NFS\cite{chen2021neural} & 78.0 & 69.8 & 80.5 & 72.1\\
    SPOT\cite{chen2022structure} & 79.4 & 72.3 & 80.4 & 72.5\\
    CM-NAS\cite{fu2021cm} & 82.6 & 78.3 & 84.5 & 80.3\\
    MPANet\cite{wu2021discover} & 82.8 & \underline{80.7} & 83.7 & \underline{80.9}\\
    SMCL\cite{wei2021syncretic} & 83.1 & 78.6 & 83.9 & 73.8\\
    DART\cite{yang2022learning} & 82.0 & 73.8 & 83.6 & 75.7\\
    CAJ\cite{ye2021channel} & 84.8 & 77.8 & 85.0 & 79.1\\
    MAUM\cite{liu2022learning} & \textbf{87.0} & \textbf{84.3} & \textbf{87.9} & \textbf{85.1}\\
    \textbf{CSL} & \underline{85.8} & 77.8 & \underline{86.2} & 77.9\\
    \bottomrule
  \end{tabular}}
  \label{tab:vi-reid-results-regdb}
\end{table}

\subsection{Comparison with CC-ReID State-of-the-Art}

\noindent Following the official evaluation metrics, performances of our CSL and all state-of-the-art methods are compared in Tab.~\ref{tab:cc-reid-results}. Our CSL outperforms all state-of-the-arts consistently.

\begin{table}[h]
  \centering
  \caption{State-of-the-art Comparison for CC-ReID on PRCC~\cite{yang2019person} and LTCC~\cite{qian2020long}.}
  \vspace{-2mm}
  \scalebox{0.85}{\begin{tabular}{@{}c||cc||cc@{}}
    \toprule
    \multirow{2}*{Methods} & \multicolumn{2}{c||}{LTCC} & \multicolumn{2}{c}{PRCC} \\ \cline{2-5}
    ~ & Rank1 & mAP & Rank1 & mAP\\ \cline{1-5}
    HACNN\cite{li2018harmonious} & 21.6 & 9.3 & 21.8 & -\\
    PCB\cite{sun2020circle} & 23.5 & 10.0 & 41.8 & -\\
    GI-ReID\cite{jin2022cloth} & 23.7 & 10.4 & 33.3 & 37.5\\
    IANet\cite{hou2019interaction} & 25.0 & 12.6 & 46.3 & -\\
    ISP\cite{zhu2020identity} & 27.8 & 11.9 & 36.6 & -\\
    3DSL\cite{chen2021learning} & 31.2 & 14.8 & - & 51.3\\
    FSAM\cite{hong2021fine} & 38.5 & 16.2 & 54.5 & -\\
    CAL\cite{gu2022clothes} & 40.1 & 18.0 & 55.2 & 55.8\\
    \textbf{CSL} & \textbf{56.2} & \textbf{22.2} & \textbf{56.4} & \textbf{56.0}\\
    \bottomrule
  \end{tabular}}
  \label{tab:cc-reid-results}
\end{table}

\subsection{Ablation Study}

\noindent Ablation studies on four benchmarks (\textit{i.e.}, SYSU-MM01~\cite{wu2017rgb}, Campus-Corridor, LTCC~\cite{qian2020long}, and PRCC~\cite{yang2019person}) are conducted in this section, as shown in Tab.~\ref{tab:ablation-study}. The effectiveness of our ICA and PCT is consistent in all datasets. Besides, our conclusion in the main paper that the ICA is essential for PCT is further proved in Campus-Campus and PRCC datasets.

To further discuss the effectiveness of ICA, we also warp our baseline with CR, CS, and Gray-scale to conduct experiments. CR and CS are discussed in the main paper, and Gray-Scale is a widely-used image augmentation method to transform the color images into gray-scale ones. The experiment results in these settings are worse than combining baseline and ICA. Specifically, CR is slightly better in VI-ReID tasks, while CS is much better for the LTCC dataset, which proves our assumption that randomly changing the image color is useful for the cloth-changing conditions. It also proves that the combination of CR and CS is meaningful, and our proposed ICA can effectively promote latent color-invariant feature learning.

We can find that after ICA and PCT, images of different colors adjust the weights of three channels (R, G, and B) automatically and dynamically adapted for different datasets (refer to Fig.~\ref{fig:pct}). For the SYSU-MM01 dataset, all PCT-transformed images become greenish (refer to Fig.~\ref{fig:pct}(a)), corresponding to the fact that sensors have double green-pixel sensors compared with red and blue sensors. For the PRCC dataset, the PCT-transformed images become yellowish (refer to Fig.~\ref{fig:pct}(b)). It accords with the knowledge that human eyes are more sensitive to green and red than blue, and the combination of green and red light should be yellowish. Our findings align with the knowledge, showing that our method can automatically learn latent information like camera sensors and human eyes.

\begin{table}[h]
  \centering
  \caption{The influence of each component of our proposed CSL approach. Experiments are conducted on both VI-ReID and CC-ReID wide-used benchmarks.}
  \scalebox{0.8}{\begin{tabular}{@{}l||cc|cc||cc@{}}
    \toprule
     \multirow{3}*{Experiment Setting} & \multicolumn{4}{c||}{SYSU-MM01} & \multicolumn{2}{c}{Campus-Corridor}\\ \cline{2-7}
     ~ & \multicolumn{2}{c|}{All-Search} & \multicolumn{2}{c||}{Indoor-Search} & \multirow{2}*{Rank1} & \multirow{2}*{mAP}\\ \cline{2-5}
     ~ & Rank1 & mAP & Rank1 & mAP\\ \cline{1-7}
     Baseline & 62.5 & 59.1 & 67.0 & 72.4 & 80.7 & 58.3\\
     Baseline+CR & 66.8 & 62.8 & 71.5 & 76.2 & 84.0 & 58.8\\
     Baseline+CS & 65.8 & 63.0 & 71.2 & 75.5 & 84.0 & 58.0\\
     Baseline+Gray-Scale & 65.3 & 62.2 & 70.2 & 74.8 & 83.6 & 57.0\\
     Baseline+ICA & 68.7 & 64.7 & 74.2 & 78.5 & 85.3 & 62.5\\
     Baseline+PCT & 65.8 & 62.5 & 70.7 & 74.8 & 83.6 & 58.9\\
     Baseline+ICA+PCT & \textbf{72.5} & \textbf{68.0} & \textbf{80.2} & \textbf{82.9} & \textbf{86.2} & \textbf{63.7}\\
     \bottomrule
  \end{tabular}}

\centering
  \begin{tabular}
  {c c}
  ~ & ~ \\
  \end{tabular}
  
\centering
  \scalebox{0.8}{\begin{tabular}{@{}l||cc||cc@{}}
    \toprule
     \multirow{2}*{Experiment Setting} & \multicolumn{2}{c||}{LTCC} & \multicolumn{2}{c}{PRCC}\\ \cline{2-5}
     ~ & Rank1 & mAP & Rank1 & mAP\\ \cline{1-5}
     Baseline & 42.3 & 15.9 & 53.7 & 52.8\\
     Baseline+CR & 42.6 & 14.6 & 54.1 & 50.1\\
     Baseline+CS & 42.6 & 14.6 & 54.1 & 50.1\\
     Baseline+Gray-Scale & 42.6 & 14.6 & 54.1 & 50.1\\
     Baseline+ICA & 46.9 & 16.6 & 54.8 & 55.1\\
     Baseline+PCT & 37.5 & 16.0 & 52.0 & 51.9\\
     Baseline+ICA+PCT & \textbf{56.2} & \textbf{22.2} & \textbf{56.4} & \textbf{56.0}\\
     \bottomrule  
  \end{tabular}}
  \label{tab:ablation-study}
\end{table}

\newpage

\bibliographystyle{ieeetr}
\bibliography{ref}

\begin{thebibliography}{10}

\bibitem{zheng2016person}
L.~Zheng, Y.~Yang, and A.~G. Hauptmann, ``Person re-identification: Past, present and future,'' {\em arXiv preprint arXiv:1610.02984}, 2016.

\bibitem{ye2021deep}
M.~Ye, J.~Shen, G.~Lin, T.~Xiang, L.~Shao, and S.~C. Hoi, ``Deep learning for person re-identification: A survey and outlook,'' {\em TPAMI}, 2021.

\bibitem{wang2019beyond}
Z.~Wang, Z.~Wang, Y.~Zheng, Y.~Wu, W.~Zeng, and S.~Satoh, ``Beyond intra-modality: A survey of heterogeneous person re-identification,'' {\em arXiv preprint arXiv:1905.10048}, 2019.

\bibitem{fan2020learning}
L.~Fan, T.~Li, R.~Fang, R.~Hristov, Y.~Yuan, and D.~Katabi, ``Learning longterm representations for person re-identification using radio signals,'' in {\em CVPR}, 2020.

\bibitem{wu2017rgb}
A.~Wu, W.-S. Zheng, H.-X. Yu, S.~Gong, and J.~Lai, ``Rgb-infrared cross-modality person re-identification,'' in {\em ICCV}, 2017.

\bibitem{yang2019person}
Q.~Yang, A.~Wu, and W.-S. Zheng, ``Person re-identification by contour sketch under moderate clothing change,'' {\em TPAMI}, 2019.

\bibitem{wang2019rgb}
G.~Wang, T.~Zhang, J.~Cheng, S.~Liu, Y.~Yang, and Z.~Hou, ``Rgb-infrared cross-modality person re-identification via joint pixel and feature alignment,'' in {\em ICCV}, 2019.

\bibitem{wang2020cross}
G.-A. Wang, T.~Zhang, Y.~Yang, J.~Cheng, J.~Chang, X.~Liang, and Z.-G. Hou, ``Cross-modality paired-images generation for rgb-infrared person re-identification,'' in {\em AAAI}, 2020.

\bibitem{park2021learning}
H.~Park, S.~Lee, J.~Lee, and B.~Ham, ``Learning by aligning: Visible-infrared person re-identification using cross-modal correspondences,'' in {\em ICCV}, 2021.

\bibitem{qian2020long}
X.~Qian, W.~Wang, L.~Zhang, F.~Zhu, Y.~Fu, T.~Xiang, Y.-G. Jiang, and X.~Xue, ``Long-term cloth-changing person re-identification,'' in {\em ACCV}, 2020.

\bibitem{gu2022clothes}
X.~Gu, H.~Chang, B.~Ma, S.~Bai, S.~Shan, and X.~Chen, ``Clothes-changing person re-identification with rgb modality only,'' in {\em CVPR}, 2022.

\bibitem{wei2021syncretic}
Z.~Wei, X.~Yang, N.~Wang, and X.~Gao, ``Syncretic modality collaborative learning for visible infrared person re-identification,'' in {\em ICCV}, 2021.

\bibitem{hu2022adversarial}
W.~Hu, B.~Liu, H.~Zeng, Y.~Hou, and H.~Hu, ``Adversarial decoupling and modality-invariant representation learning for visible-infrared person re-identification,'' {\em TCSVT}, 2022.

\bibitem{he2016deep}
K.~He, X.~Zhang, S.~Ren, and J.~Sun, ``Deep residual learning for image recognition,'' in {\em CVPR}, 2016.

\bibitem{ye2021channel}
M.~Ye, W.~Ruan, B.~Du, and M.~Z. Shou, ``Channel augmented joint learning for visible-infrared recognition,'' in {\em ICCV}, 2021.

\bibitem{cui2023dcr}
Z.~Cui, J.~Zhou, Y.~Peng, S.~Zhang, and Y.~Wang, ``Dcr-reid: Deep component reconstruction for cloth-changing person re-identification,'' {\em TCSVT}, 2023.

\bibitem{chen2021learning}
J.~Chen, X.~Jiang, F.~Wang, J.~Zhang, F.~Zheng, X.~Sun, and W.-S. Zheng, ``Learning 3d shape feature for texture-insensitive person re-identification,'' in {\em CVPR}, 2021.

\bibitem{zhang2017mixup}
H.~Zhang, M.~Cisse, Y.~N. Dauphin, and D.~Lopez-Paz, ``mixup: Beyond empirical risk minimization,'' in {\em ICLR}, 2018.

\bibitem{xia2019second}
B.~N. Xia, Y.~Gong, Y.~Zhang, and C.~Poellabauer, ``Second-order non-local attention networks for person re-identification,'' in {\em ICCV}, 2019.

\bibitem{ye2018hierarchical}
M.~Ye, X.~Lan, J.~Li, and P.~Yuen, ``Hierarchical discriminative learning for visible thermal person re-identification,'' in {\em AAAI}, 2018.

\bibitem{nguyen2017person}
D.~T. Nguyen, H.~G. Hong, K.~W. Kim, and K.~R. Park, ``Person recognition system based on a combination of body images from visible light and thermal cameras,'' {\em Sensors}, 2017.

\bibitem{zhang2022illumination}
G.~Zhang, Z.~Luo, Y.~Chen, Y.~Zheng, and W.~Lin, ``Illumination unification for person re-identification,'' {\em TCSVT}, 2022.

\bibitem{deng2009imagenet}
J.~Deng, W.~Dong, R.~Socher, L.-J. Li, K.~Li, and L.~Fei-Fei, ``Imagenet: A large-scale hierarchical image database,'' in {\em CVPR}, 2009.

\bibitem{luo2019strong}
H.~Luo, W.~Jiang, Y.~Gu, F.~Liu, X.~Liao, S.~Lai, and J.~Gu, ``A strong baseline and batch normalization neck for deep person re-identification,'' {\em TMM}, 2019.

\bibitem{liu2022learning}
J.~Liu, Y.~Sun, F.~Zhu, H.~Pei, Y.~Yang, and W.~Li, ``Learning memory-augmented unidirectional metrics for cross-modality person re-identification,'' in {\em CVPR}, 2022.

\bibitem{li2020infrared}
D.~Li, X.~Wei, X.~Hong, and Y.~Gong, ``Infrared-visible cross-modal person re-identification with an x modality,'' in {\em AAAI}, 2020.

\bibitem{hong2021fine}
P.~Hong, T.~Wu, A.~Wu, X.~Han, and W.-S. Zheng, ``Fine-grained shape-appearance mutual learning for cloth-changing person re-identification,'' in {\em CVPR}, 2021.

\bibitem{zheng2015scalable}
L.~Zheng, L.~Shen, L.~Tian, S.~Wang, J.~Wang, and Q.~Tian, ``Scalable person re-identification: A benchmark,'' in {\em ICCV}, 2015.

\bibitem{zheng2017unlabeled}
Z.~Zheng, L.~Zheng, and Y.~Yang, ``Unlabeled samples generated by gan improve the person re-identification baseline in vitro,'' in {\em ICCV}, 2017.

\bibitem{ye2018visible}
M.~Ye, Z.~Wang, X.~Lan, and P.~C. Yuen, ``Visible thermal person re-identification via dual-constrained top-ranking.,'' in {\em IJCAI}, 2018.

\bibitem{dai2018cross}
P.~Dai, R.~Ji, H.~Wang, Q.~Wu, and Y.~Huang, ``Cross-modality person re-identification with generative adversarial training.,'' in {\em IJCAI}, 2018.

\bibitem{ye2020dynamic}
M.~Ye, J.~Shen, D.~J~Crandall, L.~Shao, and J.~Luo, ``Dynamic dual-attentive aggregation learning for visible-infrared person re-identification,'' in {\em ECCV}, 2020.

\bibitem{chen2021neural}
Y.~Chen, L.~Wan, Z.~Li, Q.~Jing, and Z.~Sun, ``Neural feature search for rgb-infrared person re-identification,'' in {\em CVPR}, 2021.

\bibitem{zhao2021joint}
Z.~Zhao, B.~Liu, Q.~Chu, Y.~Lu, and N.~Yu, ``Joint color-irrelevant consistency learning and identity-aware modality adaptation for visible-infrared cross modality person re-identification,'' in {\em Proceedings of the AAAI}, 2021.

\bibitem{tian2021farewell}
X.~Tian, Z.~Zhang, S.~Lin, Y.~Qu, Y.~Xie, and L.~Ma, ``Farewell to mutual information: Variational distillation for cross-modal person re-identification,'' in {\em CVPR}, 2021.

\bibitem{lu2020cross}
Y.~Lu, Y.~Wu, B.~Liu, T.~Zhang, B.~Li, Q.~Chu, and N.~Yu, ``Cross-modality person re-identification with shared-specific feature transfer,'' in {\em CVPR}, 2020.

\bibitem{fu2021cm}
C.~Fu, Y.~Hu, X.~Wu, H.~Shi, T.~Mei, and R.~He, ``Cm-nas: Cross-modality neural architecture search for visible-infrared person re-identification,'' in {\em ICCV}, 2021.

\bibitem{chen2022structure}
C.~Chen, M.~Ye, M.~Qi, J.~Wu, J.~Jiang, and C.-W. Lin, ``Structure-aware positional transformer for visible-infrared person re-identification,'' {\em TIP}, 2022.

\bibitem{hao2021cross}
X.~Hao, S.~Zhao, M.~Ye, and J.~Shen, ``Cross-modality person re-identification via modality confusion and center aggregation,'' in {\em ICCV}, 2021.

\bibitem{zhang2022fmcnet}
Q.~Zhang, C.~Lai, J.~Liu, N.~Huang, and J.~Han, ``Fmcnet: Feature-level modality compensation for visible-infrared person re-identification,'' in {\em CVPR}, 2022.

\bibitem{yang2022learning}
M.~Yang, Z.~Huang, P.~Hu, T.~Li, J.~Lv, and X.~Peng, ``Learning with twin noisy labels for visible-infrared person re-identification,'' in {\em CVPR}, 2022.

\bibitem{wu2021discover}
Q.~Wu, P.~Dai, J.~Chen, C.-W. Lin, Y.~Wu, F.~Huang, B.~Zhong, and R.~Ji, ``Discover cross-modality nuances for visible-infrared person re-identification,'' in {\em CVPR}, 2021.

\bibitem{li2018harmonious}
W.~Li, X.~Zhu, and S.~Gong, ``Harmonious attention network for person re-identification,'' in {\em CVPR}, 2018.

\bibitem{sun2020circle}
Y.~Sun, C.~Cheng, Y.~Zhang, C.~Zhang, L.~Zheng, Z.~Wang, and Y.~Wei, ``Circle loss: A unified perspective of pair similarity optimization,'' in {\em CVPR}, 2020.

\bibitem{jin2022cloth}
X.~Jin, T.~He, K.~Zheng, Z.~Yin, X.~Shen, Z.~Huang, R.~Feng, J.~Huang, Z.~Chen, and X.-S. Hua, ``Cloth-changing person re-identification from a single image with gait prediction and regularization,'' in {\em CVPR}, 2022.

\bibitem{hou2019interaction}
R.~Hou, B.~Ma, H.~Chang, X.~Gu, S.~Shan, and X.~Chen, ``Interaction-and-aggregation network for person re-identification,'' in {\em CVPR}, 2019.

\bibitem{zhu2020identity}
K.~Zhu, H.~Guo, Z.~Liu, M.~Tang, and J.~Wang, ``Identity-guided human semantic parsing for person re-identification,'' in {\em ECCV}, 2020.

\end{thebibliography}

\end{document}